\documentclass[10pt,journal,compsoc]{IEEEtran}

\ifCLASSOPTIONcompsoc
  \usepackage[nocompress]{cite}
\else
  \usepackage{cite}
\fi

\ifCLASSINFOpdf

\else
\fi

\newcommand{\proto}[1]{{##CACTUS}}

\usepackage[T1]{fontenc}
\usepackage{txfonts}
\usepackage[pdftex]{hyperref}
\usepackage{color}
\usepackage{siunitx}
\usepackage{booktabs,arydshln}
\usepackage{textcomp}
\usepackage{multirow}
\usepackage[caption=false]{subfig}
\usepackage{microtype} 
\usepackage{ccicons}  
\usepackage{mathptmx}
\usepackage{graphicx}
\usepackage{times}
\usepackage{enumitem}
\usepackage{tabu}
\usepackage{float}

\usepackage{cite}
\makeatletter
\def\adl@drawiv#1#2#3{%
        \hskip.5\tabcolsep
        \xleaders#3{#2.5\@tempdimb #1{1}#2.5\@tempdimb}%
                #2\z@ plus1fil minus1fil\relax
        \hskip.5\tabcolsep}
\newcommand{\cdashlinelr}[1]{%
  \noalign{\vskip\aboverulesep
           \global\let\@dashdrawstore\adl@draw
           \global\let\adl@draw\adl@drawiv}
  \cdashline{#1}
  \noalign{\global\let\adl@draw\@dashdrawstore
           \vskip\belowrulesep}}
\makeatother

\begin{document}

\title{
CACTUS: Detecting and Resolving Conflicts in Objective Functions}


\author{Subhajit Das and Alex Endert

}



\IEEEtitleabstractindextext{%

\begin{abstract}
Machine learning (ML) models are constructed by expert ML practitioners using various coding languages, in which they tune and select models hyperparameters and learning algorithms for a given problem domain. They also carefully design an objective function or loss function (often with multiple objectives) that captures the desired output for a given ML task such as classification, regression, etc. 
In multi-objective optimization, conflicting objectives and constraints is a major area of concern.
In such problems, several competing objectives are seen for which no single optimal solution is found that satisfies all desired objectives simultaneously.
In the past VA systems have allowed users to interactively construct objective functions for a classifier. In this paper, we extend this line of work by prototyping a technique to visualize multi-objective objective functions either defined in a Jupyter notebook or defined using an interactive visual interface to help users to: (1) perceive and interpret complex mathematical terms in it and (2) detect and resolve conflicting objectives.
Visualization of the objective function enlightens potentially conflicting objectives that obstructs selecting correct solution(s) for the desired ML task or goal.
We also present an enumeration of potential conflicts in objective specification in multi-objective objective functions for classifier selection. Furthermore, we demonstrate our approach in a VA system that helps users in specifying meaningful objective functions to a classifier by detecting and resolving conflicting objectives and constraints.
Through a within-subject quantitative and qualitative user study, we present results showing that our technique helps users interactively specify meaningful objective functions by resolving potential conflicts for a classification task. 
\end{abstract}

}

\maketitle

\IEEEdisplaynontitleabstractindextext

%
\IEEEpeerreviewmaketitle

\ifCLASSOPTIONcompsoc

\IEEEraisesectionheading{\section{Introduction}\label{sec:introduction}}
\else
\section{Introduction}
\label{sec:introduction}
\fi

\IEEEPARstart{T}{raditionally} 
machine learning (ML) experts construct models by writing codes that includes searching for the right combination of hyperparameters and learning algorithms, and specifying an appropriate objective function (also called loss/cost functions) to the modeling task.
In the past, researchers in visual analytics (VA) have investigated making ML model construction interactive, which means developing visual interfaces that allow users to construct ML models by interacting with graphical widgets or data marks \cite{amershi2015modeltracker, clustervision}. For example, the system XClusim helps biologists to interactively cluster a specified dataset \cite{XclusSim}, Hypermoval \cite{hypermovalPir} and BEAMES \cite{beamesdas} allows interactive construction of regression models, Axissketcher allows dimension reduction using simple drag-drop interactions \cite{axiSketcher}. These VA systems use models that are driven by an objective function as designed by an expert ML practitioner or a data scientist to achieve a desired data analytic goal, such as correctly predicting class labels of unseen data, or predict a quantitative value etc. 
Recently, Das et al. have demonstrated a VA system, QUESTO \cite{questo_das} that facilitated interactive creation of objective functions to solve a classification task utilising an Auto-ML system. 
While their approach helped users to interactively explore and express a wide array of objectives to an objective function, the authors discussed potential conflicts that may occur in interactive specification of objectives as a limitation to QUESTO's workflow. For example, a user may emphasize to predict a set of relevant/critical data instances correctly, while mistakenly expressing that a subset of these data instances are outliers or noise in the data \cite{questo_das}.

	\begin{figure}[htbp]
\centering
		\includegraphics[width=3.1in]{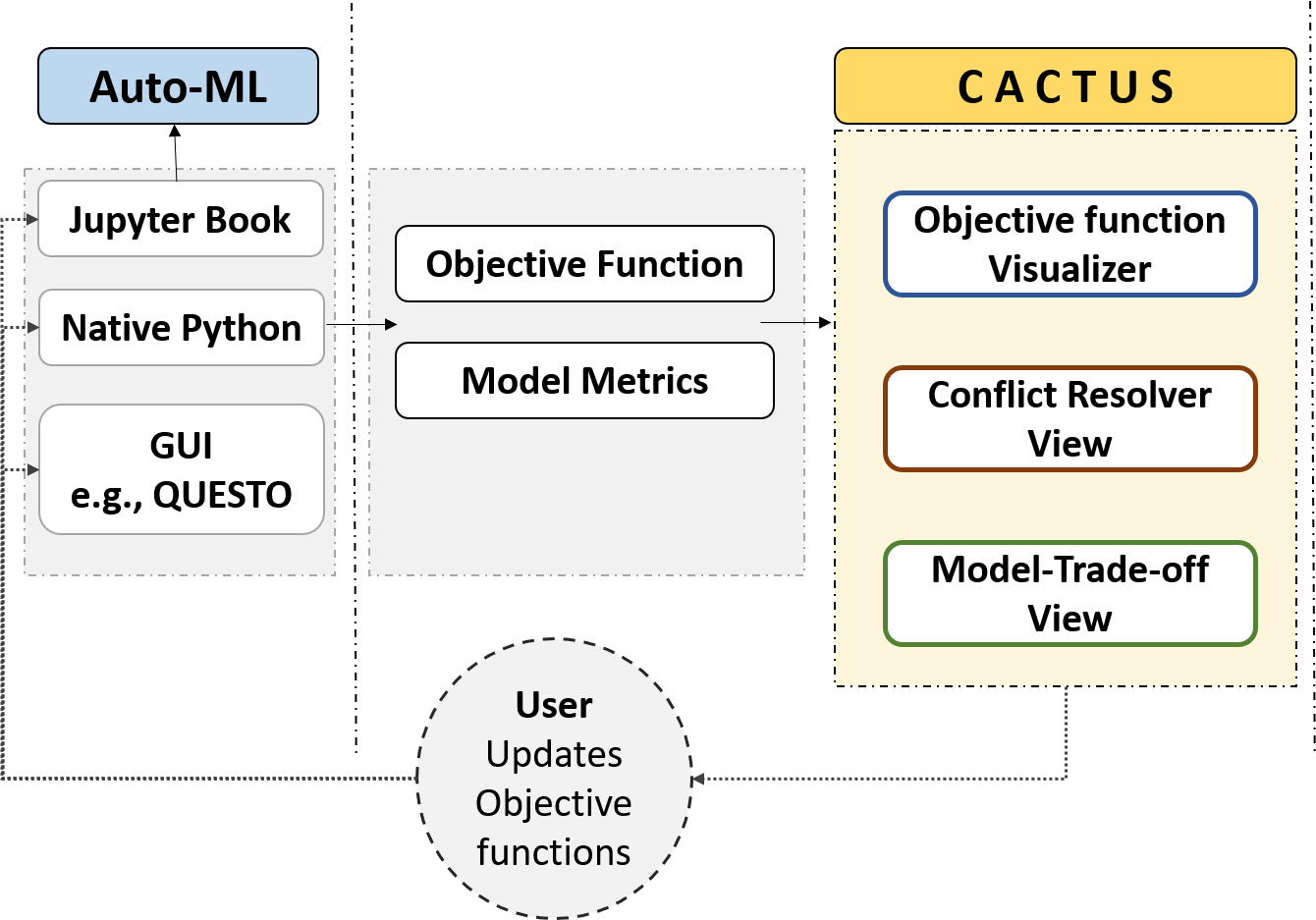}
		\caption{Workflow adopted in the system CACTUS.}
		\label{fig:cactusflow}
		\vspace{-0.3cm}
	\end{figure}
	
Interactive specification of objectives and constraints may result in objective functions with conflicting objectives \cite{NIPS2018_7334, multiobj_ML}. 
Conflicting objectives or constraints may cause construction of inefficient objective functions that may confuse the underlying algorithm (e.g., Auto-ML model solver) due to unclear user goals. 
For example, in a regression task, a user may specify to use a L2 regularizer to penalize attributes with large coefficients, but that may result into incorrectly predicting many relevant data instances, though improving the generalizability of the model. Here the objective to train a model with high accuracy on a set of important data items may conflict with the validation accuracy (or variance) of the model.
Similarly, in a classification task, a user may expect to see similar data items in the same class label; at the same time they may also expect that the global accuracy of the model is high for every class. The model being trained to support the users request to place similar data items in the same label category, may not perform equally well for all class labels, thus dropping the global accuracy of the model.
In the past conflicts in objective specification in multi-objective objective functions was addressed using trade-off analysis \cite{zhangconflict_res, conflict_obj_book, grndwater_conf, multiobj_ML}.
While useful, to our knowledge there is no other work in VA that have looked at detecting and resolving conflicts in user-specified objectives to build classification models using an interactive visual interface. 



In this paper, 
we extend research on interactive objective functions by helping users detect conflicts in objective functions, and further interactively resolving these conflicts to specify a more meaningful objective function to a model selection systems (e.g. Auto-ML).
To further understand what these conflicts are, and how adversely they may affect objective specifications, we conducted an extensive literature review.
Grounded on the literature search, we enumerated a list of potential conflicts between various objectives in interactive objective functions explained later in the paper. 
Furthermore, we present a visual analytic tool that facilitates: (1) Visualization of a multi-objective objective function that is defined using a python code (e.g, defined in a Jupyter notebook), or a visual interface such as QUESTO \cite{questo_das}, (2) Highlighting conflicts between interactively specified objectives, (3) Helping users to resolve these conflicts to improve their objective functions, and (4) Train multiple classifiers incrementally (over multiple iterations) to perform trade-off analysis of objectives in multi-objective objective function.
As objective functions drive all ML algorithms, we consider visualizing objective functions may help users (e.g., novice ML users, ML experts who may use GUI to debug models, etc.) to understand their specifications to the underlying models, and further empower them to explicitly adjust the function terms to explore and test various hypotheses that they may have. 
More importantly, we consider supporting users in defining meaningful and correct objective functions is very important to ensure models that are sampled perform as per user expectations.
With this approach, we seek to extend the current processes of interactive model tuning and model selection in which mathematical objective functions can be visualized and inspected. For example, in a Jupyter notebook one can define an objective function and then use our technique to visualize, test, and interactively adjust objectives to explore numerous model alternatives for their analytical ML task.


We prototyped CACTUS, a conflict resolution and trade-off analysis system for user specification of objectives. CACTUS ingests an objective function (defined in Python or Jupyter notebook) and then visualises it to show its objectives and respective weights. It visually explains the objectives that are closely satisfied by the selected classification model and the ones that failed to be satisfied (see Figure~\ref{fig:cactusflow}).
Furthermore, it visualises conflicts between objectives, allowing users to incrementally improve the function by making adjustments interactively. When objective functions are adjusted interactively, users can re-train models, see a change in the models' performance and continue exploring the space of multiple variants of objective functions.
Furthermore, visualising objective functions and the trained models' performance metric (per objective) can empower users to
explain/interpret and probe \cite{RibeiroSG16, Ghorbani2017InterpretationON} complex models
in relation to how well they satisfy their goals. For example, users can probe if similar data items are predicted in the same class, or strong representative (candidate) data subsets are predicted with higher probabilities by resolving conflicts and creating multiple variants of objective functions.

In addition, we present the findings from a within-subject user study. In this study we quantitatively evaluated CACTUS to test if it helped users to find and resolve conflicts, and then if it supported incremental training of classification models, facilitating comparison of a varied set of objective functions.
We also collected qualitative feedback to record user preferences in relation to easy of use, intuitiveness, and other relevant usability aspects of the system.
Our study showed that: 
(1) Participants found CACTUS intuitive and expressive in visualising complex mathematical terms in a specified objective function for a classification task. They also found CACTUS to be very effective in showing them conflicts between objectives.
(2) CACTUS helped participants ideate on multiple versions of objective functions and in the process resolve conflicts in objective specifications.
We also present qualitative feedback from the participants that enlightens the strengths and weaknesses in the current UI design of the CACTUS, potential usability issues, and limitations that needs further research in the future.
Our contributions are:
\begin{itemize}


\item An enumeration of potential conflicts in objective specification in multi-objective objective functions to construct classifiers.

\item A prototype VA system CACTUS that visualises conflicts in objective functions and supports interactive resolution of these conflicts.

\item A within-subject quantitative and qualitative user study validating that our technique helps users construct meaningful objective functions by resolving conflicts between objectives.

\end{itemize}

\section{Related Work}

\subsection{Interactive classification in VA}

In machine learning, \textit{classification} is a process to predict a class label of a data item (e.g., rows in a table) given its features in the data (e.g., columns in a table). 
During the training process, the classifier approximates a mapping function from the input data to the output class labels.
Many VA systems made this process of classifier construction interactive \cite{inmachinelearnsmalldata,convclassheir,Liu2018AnalyzingTN,deepeyesdnn}.
In some cases, this interactive classifier construction process also supports data labeling \cite{Sun17labelLearn}, hyperparameter tuning \cite{Li2018HyperTunerV}, latent space visualization of complex models such as deep neural networks \cite{Kahng2017ActiVisVE}, etc. For example, a system called ModelTracker visualized model performance (e.g., validation set's accuracy) to debug and improve models by facilitating direct data inspection by ML practitioners \cite{amershi2015modeltracker}.
Ribeiro et al. presented LIME, a prediction explanation technique that justified model results based on its performance on a set of data items \cite{RibeiroLime}. Other works that improved a users trust on ML models can be seen here \cite{Cai19effects,Yu19trust}. 
Another VA system Prospect, allowed ML practitioners to revise data properties by using multiple ML models. These models helped users understand the relationship between data features and the class label (helping users identify any noise in the class labels) \cite{Patel11MultiModel}. \looseness =-2

Furthermore, we identified few systems that are designed for intermediate users, who may know programming but are not expert in ML or data science. For example, Kim et al. showed an approach to classify sensor data that are voluminous, multi-sourced, and often unintelligible. 
They enabled pre-processing of such large sensor data and post-processing of classification model output for intermediate users \cite{makingMLsensordata}. 
Similarly, the tool Gestalt integrated ML workflows with software development for software developers (not experts in ML, but programming savvy users) who sought to interactively construct classifiers \cite{Patel10Gestalt}.
Biran et al. produced model justifications utilizing natural language generation \cite{Biran17humanjustif}.


\subsection{Interpretability in machine learning}

Conflict detection and resolution in objective functions allows users to create many variants of an objective function. In some cases, these objective function variants can act as a medium to compare models, or interpret them in relation to how they satisfy the specified objectives.
We looked at other methods in the literature, to get a sense of model interpretability in machine learning. We reckon, interactive objective functions can be another approach to interpret constructed models.
In the past, simpler models such as decision trees, linear regression, additive models \cite{letham2015interpretable, caruana15, Ustun2015SupersparseLI} were utilized to explain models as they are easier to interpret or reason about.
Compared to complex models such as deep neural networks or ensemble models, these models' internal architecture can be directly inspected and interpreted by the user. For example, users can probe various branches in a decision tree, or visualize feature weights in a linear model. 
While these models help users make sense of the prediction made by the model, its' performance accuracy falls behind compared to many state of the art complex models such as deep neural networks.
However, owing to the benefits of simpler models, recently a number of neural network architectures also utilizes interpretable components such as attention modules \cite{showattendtell15} or prototype layers \cite{prototype_18_LiEtAl18, kdd19_sequence_prototypes, prototype19_ChenEtAl2019, hase_heirarchy, icdm_proto_das} for easier human interpretability. 
Another approach to explain complex models is by using post-hoc analysis methods \cite{techforiml}.
The space of post-hoc model explanation can be divided into local and global explanation approaches further discussed below.


%

Locally interpretable approaches explain a pre-trained ML models' reasoning process with respect to input data instances.
In this space, a frequently used technique is to calculate and visualise feature attributions \cite{Raghu2017SVCCASV, unified_app_interp, smoothgrad,axiomaticdeepnetworks, Selvaraju2016GradCAMWD, Erhan2009VisualizingHF, Bach2015LRP, Shrikumar2017DeepLIFT, ancona2018towards}. Feature attributions can be computed by slightly perturbing the input features for each instance to verify how the model's prediction response varies accordingly \cite{koh2017understanding, smoothgrad}. 
Another technique in this category includes sampling features in the neighborhood of an instance to compose an additional training set. 
Using this approach an original model's prediction can be explained by an interpretable model (e.g., linear regression) that is easier to inspect \cite{RibeiroSG16}. 
A major flaw in this approach is that local explanations are shown to be less reliable and consistent as the explanations holds true only for a specific set of data instances. They cannot be extended to other similar data items in the training set.
In addition, it could also be badly impacted by adversarial perturbations \cite{Ghorbani2017InterpretationON, Kindermans2018TheO} and confirmation biases \cite{Adebayo2018SanityCF}. 
On the contrary, globally interpretable technique, focuses on explaining a models' behavior by showing a global overview, rather than showing explanation on local instances or input regions \cite{techforiml}.
For example, people have looked at interpreting latent representations learned by deep neural networks through activation maximization techniques \cite{ZeilerF13}. 
There are also concept based explanations that show how a model makes predictions globally by recovering relevant human understandable concepts \cite{kim2017interpretability, Ghorbani2019AutomatingID, yeh2019conceptbased, Zhou2018InterpretableBD}.  
In particular, concept activation vectors (CAV) are discussed by Kim et al. \cite{kim2017interpretability} as a framework to interpret latent representations in deep neural networks. 
While these works contribute to approaches on model explanation, we posit with this work we extend this further, through the creation of interactive objective functions by detecting and resolving conflicts.

\subsection{User preferences in objective functions}

We looked at the literature to understand user specifications that goes into the construction of objective functions in ML \cite{xiao18, grounding,Cen18,Cramer09} and how these functions are visualized or made interactive in the past.
User goals are often personalised and domain specific \cite{persAlz, Rudoviceaao6760}; users' may require custom-designed metrics to select models, as opposed to utilize known metrics such as `accuracy' or `recall' \cite{maniMatrixKapoor}. Often users may teach machines to show example data instances of a specific class label to help models learn how to discriminate between classes \cite{xiao18}. For instance, similarity between data instances or a sub-segment of the data is utilized to express user expectation from classifiers \cite{Tamuz11}. Similarly, in the system Flock users specified features in the data instances that rendered them similar to be in a specific class label \cite{Cheng15_flock}. \looseness =-2

Next, we searched for different ways objective functions are visualized. 
People in the past have visualized solutions recovered using multi-objective objective functions using charts such as RadViz, bubble chart, parallel coordinate plots etc. \cite{compVisObj, Sahu2011ManyObjectiveCO, Walker13}. RadViz is a popular technique used frequently to express non-dominating solutions \cite{Walker13}. A 2D-polar coordinate plot was used to help users see trade-offs between objectives in this work \cite{manyObj}. 
Others have analysed variety in pareto-optimal solutions in multi-objective optimization functions \cite{LiParCoord, diversityPareto}.
While none of these approaches allowed users to express specifications in objective functions interactively, Das et al. prototyped QUESTO, allowing users to interactively create objective functions to select optimal classifiers trained on tabular data \cite{questo_das}.
Unlike QUESTO, in this paper we seek to extend research on interactive objective functions by showing users conflicts in objectives as opposed to allowing interactive creation of objective functions.

\subsection{Conflicts in multi-objective objective functions}
Below we summarize, a set of works from different domains where people have addressed conflicts in objective specification by incorporating various learning techniques.
Real-world ML applications (e.g., finance, transportation, engineering, medical diagnosis, etc.) require addressing multiple user goals that may often conflict with one another \cite{LINDROTH20101519}. Often these goals are specified to the system using a multi-objective objective function representation \cite{coello_obj, petrie_webster_cutkosky_1995}.
In such multi-objective objective functions, it is considered as `k' (number of objectives) increases, the power of finding dominant solutions diminishes because satisfying each of the objectives becomes mathematically intractable \cite{DebOnFP}. 
Purshouse et al. confirmed that multiple objectives may be conflicted with one another; resolving conflict may show better performance in one objective than others. However, in some cases they may also show harmony that both objectives sees improved performance \cite{pursehouse_obj}. A few approaches to address better performance with a large number of conflicted objectives includes: (1) multi-start strategies of the optimization process \cite{huhes_obj}, and (2) purely random search for objective functions with more than $10$ objectives \cite{kowles_obj}.

In this context, Zhang et al. defined conflict analysis as a method to find conflicts, reason about it, and then resolve it \cite{zhangconflict_res}.
Bell et al. further described conflicts in decision making and provided an overview of quantitative approaches to address them in optimization problems \cite{conflict_obj_book}. 
Reed et al. explored scatterplot charts to visually inspect the set of conflicting objectives to solve a ground water monitoring and optimization problem \cite{grndwater_conf} (e.g., discovered a conflict between cost and uncertainty).
In machine learning, there is a recent interest in multi-objective machine learning optimization functions, which tackles conflicting user objectives. 
For example, minimizing the number of features and the maximizing feature quality are two conflicting objectives. In model selection, there is the conflict between model complexity and model accuracy (more complex, more accurate the model)
\cite{PARETO_ML, multiobj_ML}. 
Multi-task learning is another avenue where multiple tasks are solved jointly using a multi-objective optimization paradigm; however these tasks often conflict with one another that needs a trade-off analysis \cite{NIPS2018_7334}.
A common solution is to utilize a proxy objective to minimise a weighted linear combination (per task) of loss. Sener et al. showed a solution to the conflicting objectives by solving for pareto optimal solutions \cite{NIPS2018_7334}.
We observed that there are many areas in decision making, and in machine learning where conflicting objectives play a critical role. In the past, authors worked around it using elementary visualisations as a means to allow users perform trade-off analysis, where users inspect a set of pareto optimal model options. More importantly, none of the work in the past has worked towards conflict detection and resolution in objective functions, specifically designed for machine learning model selection, a problem that we are motivated to solve in this paper.

	   \begin{figure*}[htbp]
        \centering
        \includegraphics[width=\linewidth]{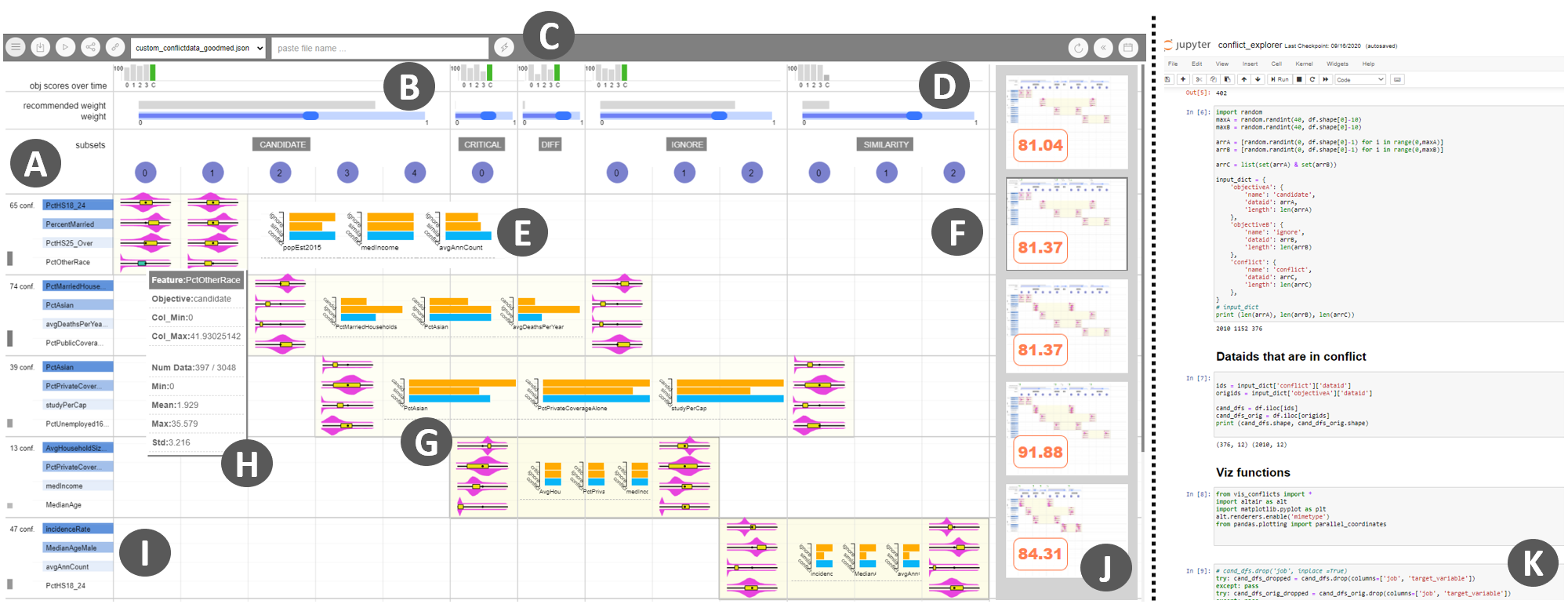}\vspace{-0.8em}
        \caption{The CACTUS system - 
        A. Data subsets per objective as seen in the header of the Conflict View.
        B. Model spark bars showing model performance per objective per iteration.
        C. Control bar to select objective functions and export objective functions.
        D. Gray bars show recommended objective weights. Blue sliders allow users to control weights per objective.
        E. Variance bars of top $3$ highly variant attributes.
        F. Conflict View.
        G. Conflict box containing the violin plots with whisker boxes.
        H. Tooltip from a whisker box showing distribution of data on an objective.
        I. Top $4$ highly variant attributes shown for each conflicts per row.
        J. Objective function gallery showing models' validation accuracy.
        K. Jupyer notebook, where objective functions can be defined.
        }
        \label{fig:teaser_CACTUS}
        \vspace{-0.3cm}
    \end{figure*}
	
\section{Types of conflicts in objective functions}

In this section we describe types of conflicts in interactive objective functions. In order to understand what these conflicts are, we adopt the objective categories as defined by Das et al. \cite{questo_das}. They defined four categories: (1) \textit{Instance-based}, (2) \textit{Feature-based}, (3) \textit{Train-objectives}, and (4) \textit{Test-objectives}. Within each of these categories there are a set of objectives such as \textit{Candidate}, \textit{Similarity}, \textit{Ignore}, etc.
Knowingly or unknowingly users may elicit numerous conflicts
while defining any of these objectives in their objective function. We categorized these conflicts in the following groups:

\begin{enumerate}

\item \noindent \textbf{Conflicts based on choices: } Conflicts can be categorized based on users' subjective choices, presuming these choices follow best ML practices (e.g., guarding against overfitting, constructing classifiers that represent every class labels precisely etc.). 

\noindent \textbf{Logic based conflicts: } 
These are conflicts that are logically incorrect but does not violate best ML practices, which we term as \textit{logic-based-conflicts}. 
For example, a user may specify a set of data items to be part of the objective
\textit{Similarity} (where these data items are expected to be in the same class label), while specifying a subset of these data items as \textit{Candidate} (an objective specifying data items that are good representative examples of a class label). These conflicts are logically incorrect but do not violate best ML practices. Refer Figure~\ref{fig:confl_matrix} to see every conflict combinations.

\noindent \textbf{ML practices-based conflicts } 
There can be conflicts which are not logically incorrect but may be violating best ML practices. For example, building a classifier using only \textit{Train-objectives}, and not \textit{Test-objectives} as a constraint, may produce overfitted models that are not generalizable to unseen data. Similarly for an imbalanced data if only an \textit{Accuracy} objective is specified as opposed to \textit{F1-Score} or \textit{Precision}. This may trigger the model solver to select classifiers that performs poorly on minority class labels. In this paper we are not addressing conflicts that may occur due to violating best ML practices. \looseness =-2

	\begin{figure}[htbp]
\centering
		\includegraphics[width=2.5in]{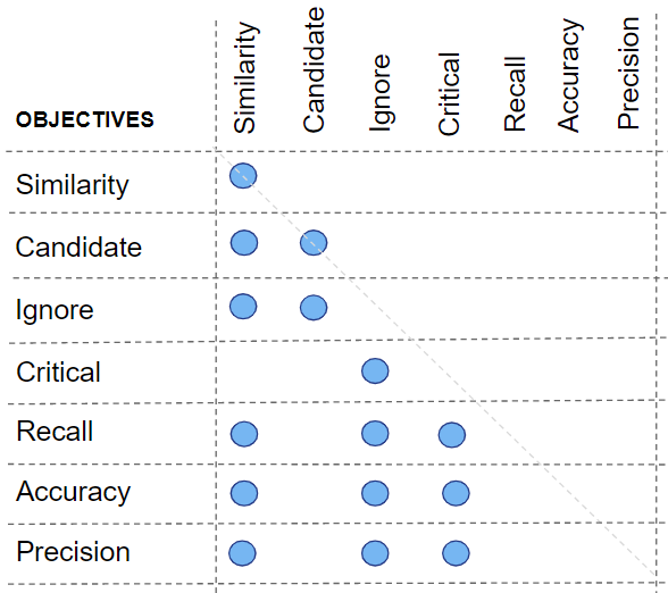}
		\caption{
		Conflict matrix, shows conflict possibilities between objectives.
		}
		\label{fig:confl_matrix}
		\vspace{-0.3cm}
	\end{figure}

\item \noindent \textbf{Conflicts based on time of occurrence: }
\textit{Logic-based-conflicts}
can be further categorized based on when they occur in the modeling pipeline.

\noindent \textbf{Before model conflicts: } 
Some conflicts can be computed before even training a model, while a few other conflicts can be only ascertained after a model is constructed. The first kind, which we term as \textit{Before-model-conflicts}, 
can be automatically computed before a ML model is constructed based on the objectives in the interactive objective function.
For example, a user has specified a set of data items - $I$ that should be placed in the same class label, say $Dog$, while also specified a subset of $I$, say $J$ data items as \textit{Candidate} objective for the class label $Cat$. Here \textit{Candidate} objective means data items that are strong representative of a specified class label. So the conflict is that the same data items are specified as examples of two different label categories $Dog$ and $Cat$. 
Similarly another example of a \textit{Before-model-conflict} is between the constraints \textit{Critical} and \textit{Ignore}. While \textit{Critical} represents the data items that are very important for the user, and thus the user expects the model to predict them correctly, while \textit{Ignore} represents data items that are unimportant, or noise or garbage in the training set. A user may specify a set of data items as \textit{Critical}, while in a future iteration of the model construction may specify a subset of \textit{Critical} data items as \textit{Ignore}. Such kinds of conflicts can be computed before a model is constructed.

\noindent \textbf{After model conflicts: }
In addition to the above, there may be other types of conflicts that are only noticed when a model is constructed or many iterations of model construction have occurred which we term as \textit{After-model-conflicts}. From the literature we know that in a multi-objective objective function, all objectives cannot be attained \cite{multiobj_ML}. In such scenarios, a set of pareto-optimal solutions are presented to the user in which only a subset of objectives are attained in each of the pareto-optimal solutions \cite{manyObj}. Analysing model log data over multiple modeling iterations, the system can infer which objectives are repeatedly unattained, or which set of objectives cannot be solved together (at the same time). In such cases, these objectives are in conflict with one another, meaning that the specification of one, blocks attainment of the other or vice versa in the objective function \cite{zhangconflict_res}. For example, the system may infer that a highly weighted \textit{Train-Accuracy} objective is prohibiting the model solver to find a model that also attains the \textit{Similarity} constraint successfully. These conflicts can only be inferred when the model is constructed or when multiple iterations of model construction has occurred. 

\end{enumerate}
 
In this paper, we scope our conflict detection and resolution method to handle conflicts that are logically incorrect and of the type \textit{Before-model-conflict}, but do not violate best ML practices. 

\section{Design Guidelines and Tasks}
We have formulated the following design guidelines for CACTUS:

\noindent \textbf{DG1: Visualize the objectives and the objective function. } We seek to build a system that is able to visualize an objective function, showing its component objectives and assigned weights. Users should be able to visually interact with the function and understand what it means to the underlying model solver where it is ingested to sample classifiers.

\noindent \textbf{DG2: Visualize model performance on each objective. }
Through our system we seek to show performance of the selected model on each of the specified objectives for every iteration of model construction. Users should be able to visually perceive how the model satisfied the specified objectives.

\noindent \textbf{DG3: Show conflicts in various objectives. }
The system should visually show conflicts between objectives. Users should be able to visually understand the conflicts and seek details on each conflict (e.g., distribution of the conflicted data, see variance of attributes etc.).

\noindent \textbf{DG4: Resolve conflicts: }
The system should help users interactively resolve conflicts. It should help them understand the conflicts in relation to the data items.
Furthermore, the system should provide affordances to either perform conflict resolution through CACTUS or allow users to export conflicts to a Jupyter notebook (where users redefine objectives to improve the function).


Based on the above guidelines we set forth the following tasks that CACTUS should support:

\noindent \textbf{T1:}
Import and visualize objective functions from a Jupyter notebook (NB). In addition, allow users to switch back and forth between multiple objective functions that they import throughout their analysis process.

\noindent \textbf{T2:}
Inspect conflicts between objectives
to address potential problems with the objective function. Be able to compare severity of conflicts, between any objectives in a function, and between multiple objective functions.

\noindent \textbf{T3:}
Resolve conflicts by re-assigning conflicted data items to a chosen objective. They should also be able to control implication of any conflict by adjusting weights assigned to these objectives. Users should be able to export conflicted data items' ID or the entire objective function to NB to further redefine the objective. 

\noindent \textbf{T4:}
Incrementally construct many ML models and inspect model trade-offs between multiple model alternatives, through the interactive creation of objective functions. These functions can be variants of the function that they import created in the process of conflict resolution, re-assigning weights or re-writing the function in NB and importing again.

\section{CACTUS: System Design}
Here we describe the user interface and interactions supported by CACTUS. The main views of the system are: (1) Conflict view, (2) Model spark bars, (3) Venn diagram view, (4) Feature plots, and (4) Objective functon gallery.

\subsection{User Interface}

\begin{figure}[htbp]
\centering
		\includegraphics[width=3.4in]{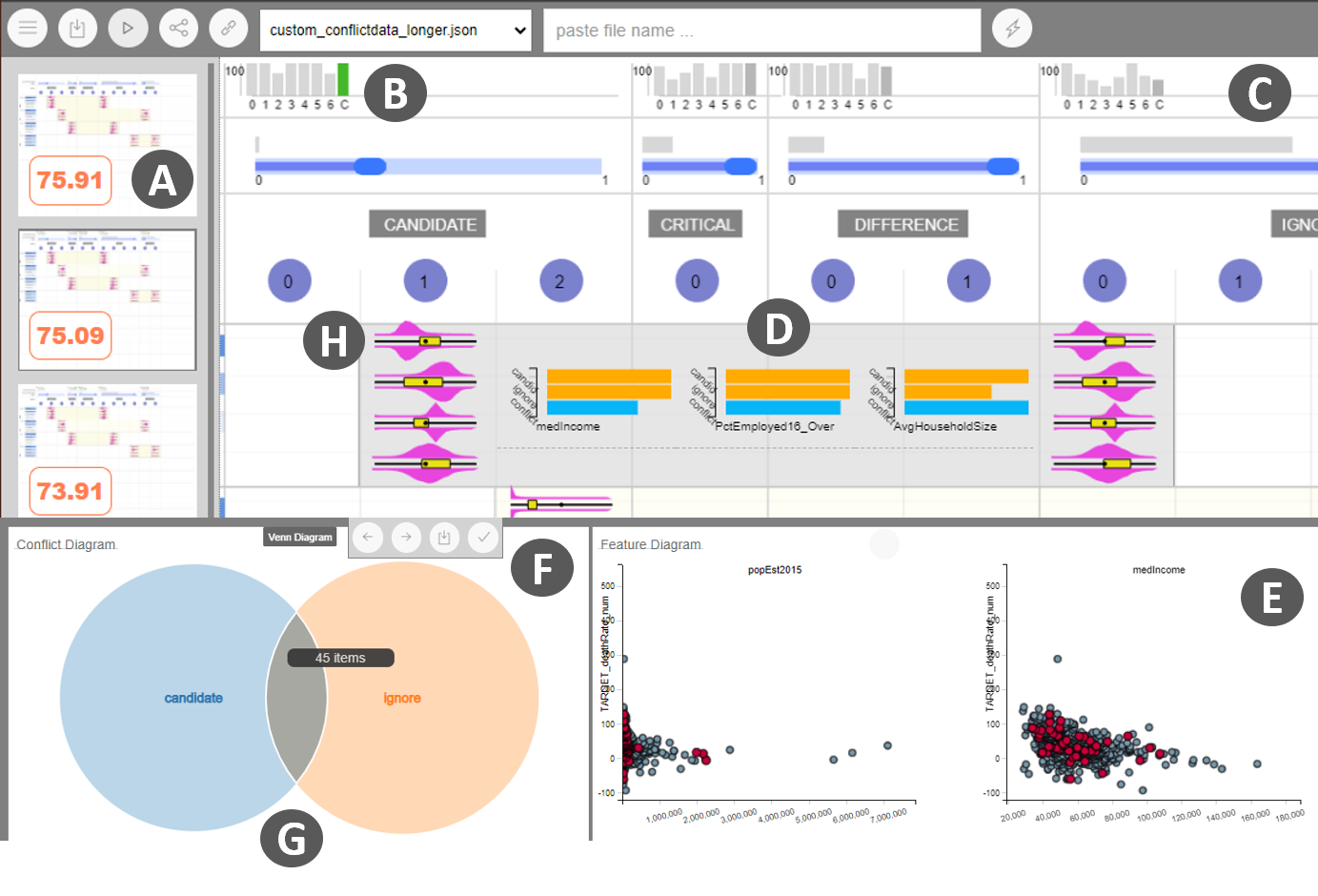}
		\caption{
        A. Objective function gallery shows previews of objective function per iteration with model validation accuracy.
        B. Model spark bars, green bars reflect current iterations performance improved over previous iteration.
        C. Gray bars show recommended weights. Blue bars are sliders to allow users to specify weights to objectives.
        D. Variance bars to compare conflicted data items' distribution with other objectives.
        E. Feature plots, blue dots are full training set, red dots are objective or conflicted data items.
        F. Context menu to resolve conflicts.
        G. Venn diagram view showing conflicted data overlap between objectives.
        H. Conflict box opens the bottom drawer to show venn diagrams and feature plots. \looseness =-2
		}
		\label{fig:venn_feature}
		\vspace{-0.3cm}
	\end{figure}

\begin{figure}[htbp]
\centering
		\includegraphics[width=3.2in]{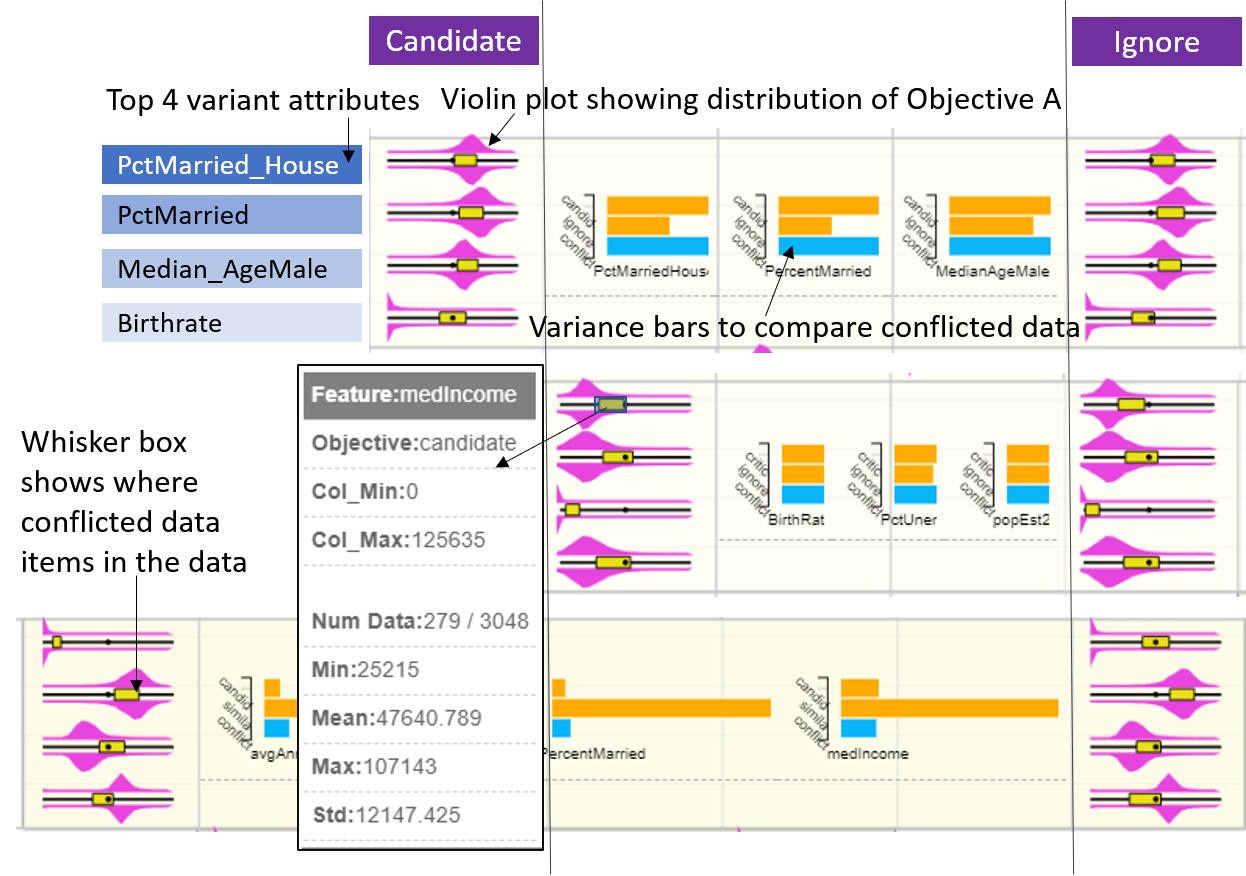}
		\caption{
		Violin plots and whisker boxes show how conflicted data items are similar or different to the pair of objectives.
		}
		\label{fig:violinplots}
		\vspace{-0.3cm}
	\end{figure}

\noindent \textbf{Conflict view: }
This view shows conflicts using a table representation, where every column shows an objective from the loaded objective function (see Figure~\ref{fig:teaser_CACTUS}-F, \textbf{DG1}). The second row (with the numbered circles) represent subset data instances that were specified as examples as part of the respective objective. 
Next, every row in the table encodes a conflict (\textbf{DG3}) between a pair of objectives (e.g, \textit{Similarity}, and \textit{Ignore}).
Conflict pairs are emphasized by a highlighted rectangular box (also called \textit{Conflict box}, see Figure~\ref{fig:venn_feature}-H), where $4$ most highly variant attributes are vertically ordered. 
Aligned with each of these attributes, 
a violin plot, with a whisker box plot is rendered (Figure~\ref{fig:teaser_CACTUS}-G and Figure~\ref{fig:violinplots}).
Here the violin plots show the distribution of the data in relation to that attribute, and the whisker plot shows the shape of the data instances that are part of the objective. \looseness = -2

\begin{figure}[htbp]
\centering
		\includegraphics[width=3in]{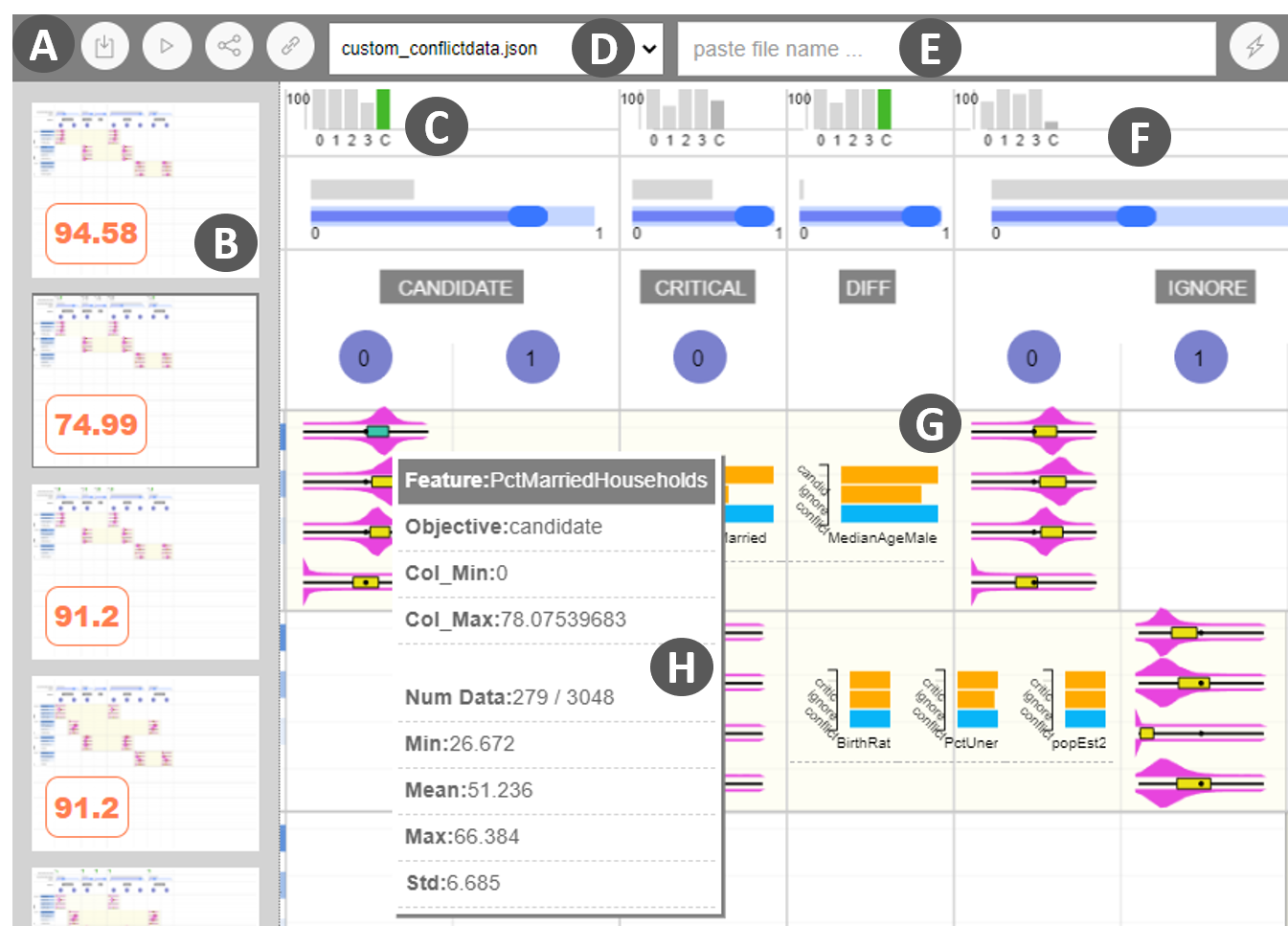}
		\caption{
		A. Buttons to train models, load and generate objective functions.
        B. Objective function gallery.
        C. Model spark bars encoding objective satisfaction score.
        D. Drop down objective function selector.
        E. Specify local path to an objective function.
        F. Objective weights.
        G. Conflict box and Variance bars.
        H. Tooltip seen on mouse hover on whisker boxes.
		}
		\label{fig:objfunc_gal}
		\vspace{-0.3cm}
	\end{figure}

Further, users can hover their mouse on the whisker box to see detailed information about the shape of the data (Figure~\ref{fig:teaser_CACTUS}-H and \ref{fig:objfunc_gal}-H).
Thus, using this visual technique, users can compare the conflicted data instance's shape across two objectives that are in conflict with one another.
The Conflict box, also shows a horizontal bar chart of \textit{Variance bars}, where it shows the variance of the data in the two objectives (shown in orange) and the variance of the conflicted data items (shown in blue) in relation to top $3$ highly variant attributes (see Figure~\ref{fig:venn_feature}-G). Using this view, users can understand how similar or different are the conflicted data in comparison to the data items that are part of the two objectives. 

\noindent \textbf{Model spark bars: }
On top of the Conflict view, CACTUS also renders horizontal sliders that allow users to interactively specify weights to the objectives (\textbf{DG2}). In addition, users can refer to system recommended weights, if they are unsure about the weights for each objective (Figure~\ref{fig:teaser_CACTUS}-D). CACTUS allows users to incrementally train models while they adjust the objective function (e.g., by resolving conflicts or updating objective weights). Per iteration when a new model is trained, its validation accuracy is plotted as a series of vertical bars, called \textit{Model spark bars}, scaled between $0$ to $100$ as seen in Figure~\ref{fig:teaser_CACTUS}-B). These accuracies are computed only on the objective of the column they are rendered in, thus allowing users to compare performance of the model with respect to specific objectives.

\noindent \textbf{Venn diagram view: }
As users explore the conflicts, they can click on the \textit{Conflict box} between an objective pair, to trigger the system to open the bottom tray.
It reveals a venn diagram showing the overlap between the pair of objectives. The size of the circles reflect how many examples comprise each objective, while the overlap of the circle encodes the conflicted data items (see Figure~\ref{fig:venn_feature}-G) that are shared by both objectives.
Furthermore, users can hover over the overlapped region to see the distribution of the data on the \textit{Feature plots} (Figure~\ref{fig:venn_feature}-E) and also on the \textit{Variance bars} (see Figure~\ref{fig:venn_feature}-D and Figure~\ref{fig:violinplots}).
Based on their exploration, they may decide to resolve the conflicts (\textbf{DG4}) by either: (1) Moving the points to the left or to the right objective, (2) Exporting the conflicted data to the Jupyter notebook, and (3) Completely removing the conflicts from the objective function using the context menu seen in Figure~\ref{fig:venn_feature}-F.
Resolving any conflict, removes the visual representation of it, i.e., a single row from the \textit{Conflict view} is removed from the display. \looseness = -2

\noindent \textbf{Feature plots: }
This view plots a set of $k$ attributes with highest variance as small multiples of scatterplots (Figure~\ref{fig:venn_feature}-E). Each of these charts are plotted with the target variable (or dependent variable) on the y axis. The design of this view is intended to show users the shape of the data in an objective or in a conflict in relation to the input data (\textbf{DG4}). This view is linked with the \textit{Venn diagram view}. \looseness =-2

\noindent \textbf{Objective function gallery: }
As users resolve conflicts or change weights of objectives, a new version of the objective function is stored in the memory. 
Users can access the history of objective function creation through this view, where each state of the objective function is shown using a thumbnail preview (see Figure~\ref{fig:teaser_CACTUS}-J). The preview also shows the trained models' validation accuracy score scaled between $0$ to $100$ (\textbf{DG2}). This view also supports users to go back to a previous state of the function, (e.g., to use a function with a better model performance score).

\subsection{Conflict Detection and Resolution Method}
Next we explain how CACTUS detects conflicts and visualizes objective functions in relation to the found conflicts.

\noindent \textbf{Conflict Parser: }
A user may write an objective function, $O$ in Python/Jupyter notebook comprising of a set of $k$ objectives $\omega_{1}, \omega_{2}, \omega_{3}, ....  \omega_{k},$. Furthermore, each of them can be specified with a set of $k$ scores ($s_{1}, s_{2}, s_{3}, ....  s_{k}$). Thus, $O$ can be represented as a weighted linear combination of these objectives as seen here,
$O = s_{1}* \omega_{1} + s_{2} * \omega_{2} + s_{3}* \omega_{3} + s_{4}* \omega_{4} + s_{5}* \omega_{5}$.
An objective $\omega_{i}$ is represented as a set of training data instance $T$ ID's as $t_{1}, t_{2}, t_{3}  .... t_{l}$. However, in the case of \textit{Candidate} objective, they may also be stored as ID's for a specific  class label $L_{1}$, $L_{2}$,...$L_{b}$, ($b$ class labels).
The conflict parser module of CACTUS checks for any overlap between all the paired combination of objectives from $O$. For example, it utilizes $T_{i}$ and $T_{j}$ between the  objectives $\omega_{i}, \omega_{j}$, using the function $FN(\omega_{i}, \omega_{j})$ to find conflicted data ID's $T_{c}$.
Finally, this module generates a hashmap object $F$, where keys are hashed to represent the objective pairs ($\omega_{i}-to-\omega_{j}$), and the values are the conflicted data ID's $T_{c}$. \looseness = -2

\noindent \textbf{Data Distribution and Variance: }
Conflicts are visualized using the $F$ object (generated by the \textit{conflict parser} module). Sequentially, the system tracks the pair of objectives $P_{i}$ that are in conflict using the hash-keys ($f_{k}$) of the object $F$. Next, it retrieves the set of data ID's $T_{i}$ that are specified as examples as part of the objective pairs $P_{i}$. It also recovers $T_{j}$ from $F$ that represents the conflicted data items. Using $T_{i}$, the system first retrieves the top $3$ (this number can be adjusted) attributes with highest variance in this set. Next it draws the violin plot $V$ and the whisker plot $W$. While $V$ shows the distribution of the full training set, the whisker plot $W$, shows the distribution of the examples that are part of the respective objectives in $P_{i}$. 
Similarly, the \textit{Variance bars}, are rendered to visualise the variance for the data ID's in $P_{i}$, and the conflicted data items $T_{j}$. The venn diagrams are also drawn using the object $F$.

\noindent \textbf{Model Solver: }
Users can also utilize any Auto-ML model solver $M$ to sample $n$ classifiers $C$ (e.g., can be adjusted $n=200$) in Python or in a Jupyter notebook.
For the current prototype we tested with Hyperopt \cite{komer2014hyperopt}, and Optuna \cite{optuna_2019} but can be replaced by Auto-SKLearn \cite{autosklearn}, or Auto-Pytorch \cite{zimmer_autopytorch} if required. In this pipeline, $M$ expects an objective function $O$, to score each of the classifiers $c_{i}$ in $C$. The highest scoring ($H$) classifier $c_{k}$ is selected and its'
overall performance (e.g., validation accuracy score), and per objective performance is visualized in CACTUS.
While the model construction part is handled by Python or the Jupyter notebook, CACTUS only ingests the objective function $O$, and visualizes conflicts.

\noindent \textbf{Weight Recommendation: }
The set of weights $S = s_{1}, s_{2}, s_{1} .... s_{k}$ in $O$ can be specified from the Jupyter notebook. These can also be interactively adjusted from the interface of CACTUS using the sliders (Figure~\ref{fig:venn_feature}-C). To further guide users, our approach also recommends weights $S'$ (between $0$ and $1$, for each objective $\omega_{i}$). 
In the initial iterations, the recommendations are randomly initialized, however, as users incrementally construct multiple versions of objective functions ($O = O_{1}, O_{2}, O_{3}, ... O_{f}$) and models ($M = M_{1}, M_{2}, M_{3}, ... M_{f}$), these weights are recommended based on the probabilistic likelihood of the weight settings that found success in: (1) Maximising the overall model accuracy, and (2) Maximising the objective score for which the weight is recommended. 
We followed the Bayesian approach \cite{autosklearn} in modeling the probabilistic likelihood of the weight settings.

\section{Case Study}

Here we describe a case study, where a user works in Jupyter notebook (NB) along with CACTUS to construct a classifier. In this process, they design multiple variants of an objective function and then visually inspect their effectiveness. 
Using CACTUS's conflict resolution technique, they discover and resolve conflicts in the specified objectives in the objective function.
Consider James is a data analyst in the public policy department of US Government and seeks to construct a classifier to predict \textit{Cancer mortality rate} in the country.
They have access to a \textit{Cancer mortality dataset (per US County)} \cite{cancerdata} containing $3048$ records (rows in the table);
each row represents a US county. It contains $34$ independent variables such as
\textit{incident-rate}, \textit{median-age}, \textit{avg-household-size}, \textit{birth-rate}, \textit{perc-resid-health-coverage}, and others.
James wants to predict the \textit{Death-Rate-Per-County}, annotated by the labels: \textit{negligible}, \textit{low}, \textit{medium}, \textit{high}, and \textit{very-high}.

\vspace{.1cm}
\noindent \textbf{Define and visualize objective functions: }
James uses Jupyter notebook (NB) to explore and analyse the data (Figure~\ref{fig:teaser_CACTUS}-K). To construct a classifier they first partition the input data $U$ into three sets: (1) Training set $R$ ($2300$ samples), (2) Validation set $S$ ($600$ samples), and (3) Test set $T$ ($100$ samples). They further partition the validation set into multiple subsets. 
Next, James writes a Python class object to construct a gradient boosted model
to classify the data. When trained using this model, James observes a relatively poor accuracy of only $72\%$, and $65\%$ on the training and validation set respectively. Motivated to select a preferred optimal model for this problem, James decides to use Optuna - a hyperparameter tuning/Auto-ML solver  \cite{optuna_2019}, and writes a custom objective function for this package to solve for.
James defines an objective function in NB containing a set of objectives including \textit{Candidate}, \textit{Ignore}, and \textit{Critical}.
At this point James re-trains a new classifier by feeding this objective function in Optuna, and notices that the training accuracy improved to $86\%$, while the validation set improved only marginally ($69\%$ accuracy). In their objective function specification James filters the data with less than value $50000$ of the attribute \textit{med-income} and higher than $18 \%$ of the attribute \textit{poverty-percent} to be classified in the same class label of \textit{medium}, as they find them to be similar to one another. Furthermore, they select a set of data samples $F$ whose $birth-rate$ and $median-age$ attribute values range between $6$ to $15 \%$ and $35$ to $50$ respectively, as critical counties that should be correctly classified. Thus in the objective function, James specifies a condition to penalize any sampled model if it makes a mistake in predicting any of the data items in $F$. \looseness = -2

\vspace{.1cm}
\noindent \textbf{Discover and remove conflicts in objectives: }
Motivated to explore different variations of this objective function and to validate any conflicts between the specified objectives, James imports it in CACTUS (on a separate tab in the browser, Figure~\ref{fig:teaser_CACTUS}). 
James sees the conflicts in the \textit{Conflict view}, and 
looks at the left end of the row to find the most severe conflict (Figure~\ref{fig:teaser_CACTUS}-A).
They see that there are $39$ data samples that are in conflict between the objectives \textit{Similarity} and \textit{Candidate}.
They also inspect the conflicted data items' position with respect to the attribute values shown through the violin plots and whisker boxes as seen in Figure~\ref{fig:violinplots}.
James notices that the examples shown as part of the objectives have similar shape and thus likely to be confusing the model solver.
They click on its \textit{Conflict box} to reveal the \textit{Venn diagram view} (see Figure~\ref{fig:venn_feature}-G) showing the conflicted data samples. 
Next to this view, they explore the \textit{Feature plots} to see the data points in each of these objectives, along with the ones that are conflicted (Figure\ref{fig:venn_feature}-E). 
Hovering over each of the venn diagram's arc sectors, James inspects the data samples on the \textit{Feature plots} to answer if these samples are similar or different from the two objectives (as seen in the red dots in Figure~\ref{fig:venn_feature}-E). 
James realizes that a condition to specify a set of data samples with attribute values of \textit{med-income} lesser than $50000$, might have caused this conflict. 
They export this conflict to Jupyter notebook (NB) to further re-define the objectives \textit{Similarity}, and \textit{Candidate}.
In NB James adds two more objective subset examples for: (1) \textit{Similarity}, and (2) \textit{Critical}. For the \textit{Candidate} objective, James specifies a set of exemplary counties that are good examples of the class label \textit{high} for death-rate prediction.
James trains a new model and exports the new objective function to CACTUS for further analysis. \looseness = -2

\vspace{.1cm}
\noindent \textbf{Iterative modeling by conflict resolution: }
CACTUS updates the \textit{Conflict view} with the new objective function. James looks at another conflict between \textit{Candidate} and \textit{Ignore} (see Figure~\ref{fig:teaser_CACTUS}). They click on its \textit{Conflict box} to inspect the venn diagram.
They understand that $74$ examples specified in the class \textit{medium} for \textit{Candidate} are actually from different class labels.
In addition, they see a subset of the \textit{Candidate} examples are also specified as part of the \textit{Ignore} objective.
To test the models' performance without a conflict between \textit{Candidate} and \textit{Ignore}, they reduce the weight on the objective \textit{Candidate} to $0$. 
James trains a new model using this weight setting (Figure~\ref{fig:venn_feature}-C).
CACTUS visualizes the newly trained classifiers' metrics on the \textit{Model spark bars} on top (Figure~\ref{fig:venn_feature}-B).
and overall models' validation set accuracy in the \textit{objective function gallery}.
James notices the model attained a \textit{Validation-accuracy} score of $80.1\%$ (Figure~\ref{fig:teaser_CACTUS}-J); they understand that by removing the objective \textit{Candidate} (by setting its weight to $0$), marginally helped the models' performance. \looseness = -2


Next, James seeks to improve the classifiers performance further. They first export the objective function and the conflicts to NB (using the buttons seen in Figure~\ref{fig:objfunc_gal}-A) and then inspect the model to see if there are any discrepancies in the specified examples.
They find many important counties are incorrectly classified by this model. James re-defines the \textit{Similarity} and the \textit{Candidate} objective to take note of it. 
From CACTUS James observes that one of the most severe conflict was between \textit{Critical} and \textit{Ignore}.
To remove the conflict between them James filters the conflicted data samples using the data ID's retrieved from CACTUS. They find similar data samples (using cosine distance) to the conflicted data samples from the training set for the \textit{Critical} objective. They replace the similar data samples in the \textit{Critical} objective to resolve the conflict with the \textit{Ignore} objective.
Next, James trains a new classifier, loads it in CACTUS to see that the \textit{Validation-set} accuracy improved to $94.58\%$ (Figure~\ref{fig:objfunc_gal}-B). 
Happy with the analysis results and the models' trained James exports the best model and the final objective function to share with their collaborators.
In this usage scenario, James uses Jupyter notebook and CACTUS hand-in-hand, to design objectives and remove conflicts respectively. The workflow helped them select a classifier that performs optimally on specified objectives. \looseness = -2


\section{Evaluation}
We conducted a qualitative and quantitative user study of CACTUS to validate the effectiveness of our conflict resolution technique. As we did not find any other visual analytic system that helps users to find and resolve conflicts in interactive objective functions, we could not design our study to compare results and prove statistical significance of any measure. Thus, given the constraints, we designed our study to address the following research questions:


\begin{description}
    \item[RQ1] Does CACTUS make it easy to precisely find conflicts between objectives in objective functions for classifiers?
    \item[RQ2] Does CACTUS support users in correctly resolving conflicts between objectives?
    \item[RQ3] Does CACTUS help users to compare objective functions and understand trade-offs between them?
\end{description}


We recruited $14$ participants ($9$ Male, $5$ Female), 
aged between $22-36$ ($M=26.06 \ [22.41, 29.71]$), by inviting participants through our university mailing lists. Our requirement was that they should know how to read/write basic python code, with elementary understanding of classifier construction and exploratory data analysis. Our participants were a mix of Masters and PhD students from computer science, analytics, geography, and urban planning. They had basic familiarity with data analysis ($M=5.26 \ [3.73, 6.79]$, on a Likert scale rating of $1-7$, higher is better), and basic ML expertise ($M=4.85 \ [3.63, 6.07]$).
The study was conducted remotely using Bluejeans \footnote{https://www.bluejeans.com/} in light of the on-going COVID-19 pandemic. It lasted $60$-$70$ minutes and at the end of a successful session we compensated participants with a \$$10$ Amazon gift card. The system was deployed on our computer, which we shared using a publicly accessible URL retrieved using NGROK \footnote{https://ngrok.com/}, with participants to conduct the study.

\subsection{Study Design}

We began the study by using a live demo,  showing participants how CACTUS works and how its' visualizations can be interacted with.
We also demonstrated how the system integrated with a Jupyter notebook environment to seek objective functions (pre-defined by writing Python scripts), and data instances.
During this session, we encouraged participants to ask as many questions they wanted to clarify any confusion with respect to the workflow or the system interface. 
Next, when we felt confident that participants were ready for the tasks, we prcoceeded to the experimental sessions.
To answer the previously mentioned \textbf{RQ}'s we considered these dependent variables: (1) \textit{Task completion times} to detect/find and resolve conflicts, (2) \textit{Conflict resolution success rate}, i.e, the number of conflicts the participants correctly resolved out of the total conflicts for all the given objective functions (between $0$ to $1$) etc., (3) \textit{Model Accuracies}, accuracy score of models per iteration of objective function specification (scaled between $0$ to $1$), (4) Number of iterations as users incrementally created objective functions by resolving conflicts, and (5) \textit{User preference ratings} that includes \textit{Ease of use}, \textit{Intuitiveness of the GUI}, \textit{Steep learning curve}, and other relevant system interactions (all of the scores were normalized between $0$ to $1$).


\subsection{Datasets}
For the practice session, we provided a dataset of $5000$ IMDB movie records \cite{moviesData}. The data had attributes such as \textit{gross-revenue}, \textit{budget}, \textit{cast-facebook-likes}, \textit{number-user-votes}, etc. It was a multi-class classification task to predict the rating of a movie between \textit{low}, \textit{moderate}, \textit{high}, and \textit{very-high}.
For the first experimental session, we provided San Francisco city's employment dataset \cite{salData} containing $25000$ records of job types for the quantitative evaluation. Each row in the data contained information about a job's remuneration information containing attributes such as  \textit{dental-benefits}, \textit{annual-salary}, \textit{health-benefits}, \textit{retirement-compensation}, etc. The task was to predict the job's department which had $5$ classes e.g., \textit{Cultural/Recreation}, \textit{Public Service}, \textit{Healthcare}, \textit{Administration}, and \textit{Other}. 
For the next experimental session, we provided the \textit{Cancer mortality dataset (per US County)} \cite{cancerdata} to predict \textit{Death-Rate-Per-County}. The class label were {Very high}, \textit{High}, \textit{Moderate}, \textit{Low}, and \textit{Negligible}.
The dataset contained $3048$ rows, each row representing the death rate of a US county and was annotated with one of the five categories of class labels. 
Furthermore, it had $34$ attributes ($1$ categorical variable) including \textit{incident-rate}, \textit{median-age}, \textit{avg-household-size}, \textit{birth-rate}, \textit{perc-resid-health-coverage}, and others.

\subsection{Tasks and Procedure}

In the practice session we provided users with a list of $3$ pre-defined objective functions (written in a Jupyter notebook) on the IMDB movies dataset \cite{moviesData}. We asked them to load each of the objective function in CACTUS and visually explore the conflicts in various objectives. Next after $15$ minutes of practice we asked them questions such as: (1) Which objective pair has the highest conflict? , (2) Name the top $2$ highly variant attributes for the conflict between \textit{Ignore} and \textit{Similarity}, (3) Resolve conflicts between \textit{Similarity} and \textit{Candidate}. (4) Export conflicts between \textit{Candidate} and \textit{Ignore} to the Jupyter notebook, and (5) Adjust the weights of the objectives and train a new model on each of the given objective functions, and then export the best objective function based on model accuracy score. 
When we ensured they understood the concept and the interactions supported by the system, we moved on to the experimental sessions.
For session $A$ we asked participants to load three objective functions (pre-defined by us in a Jupyter notebook) on the San Francisco's salary dataset \cite{salData}. We asked them to perform the following tasks:

\begin{description}
    \item [Task 1] Report number of data items that are in conflict between \textit{Ignore} and \textit{Candidate} from the second objective function.,
    
    \item[Task 2] Name the top $2$ attributes with high variance between the objectives \textit{Similarity} and \textit{Candidate}.
    
    \item[Task 3] Resolve the first and the last conflict from the third objective function. Train a new model after you resolve each conflict and then compare the model performance. Export the objective function with better model accuracy.
    
    \item[Task 4] Out of the three objective functions, find the objective function that has the highest conflict between any of the objectives.
    
    \item[Task 5] Train three models by changing weights of any of the objectives for each of the three objective functions. Which objective function found the better performing model?
    
    \item[Task 6]  Export any conflict from the the top $2$ best performing objective function from your list of saved objective functions, to Jupyter notebook.

\end{description}

In the next session (B) we asked participants to freely use CACTUS using a given pre-defined objective function and a given baseline classifer on the Cancer mortality dataset \cite{cancerdata}. Their task was to:

\begin{description}
    \item[Task 7] Incrementally improve  the baseline models’ accuracy using any of the interactions supported in CACTUS in $8$ minutes.
\end{description}

In total participants performed $7$ tasks using $2$ datasets to build a set of classifiers. To remove any learning effect, we randomised the tasks across participants.

\subsection{Data Collection}

We captured video and audio of participants screen while they interacted with CACTUS.
We saved log data containing models' selected by users per iteration. It also included models' learning algorithms, and hyperparameters, predicted class labels, interacted objectives and conflicts, etc. 
When participants completed all the tasks for both sessions, we asked them to fill a NASA-TLX form \cite{nasatlx}, and a post-study questionnaire with a set of Likert scale questions (in a 7 point scale).
In the end we conducted a semi-structured interview asking open-ended questions about the workflow, system usability, and interaction design of the interface.
For example we asked: (1) Explain your strategy to resolve conflicts? (2) Elaborate your thoughts on CACTUS's workflow, design and interactions. (3) How can we improve the current design of CACTUS?
Throughout the tasks we encouraged participants to think aloud while they interacted with CACTUS. 
Next we present quantitative and qualitative results from the study.

\subsection{Quantitative Analysis}
We broadly measured if using CACTUS: (1) users can detect conflict easily and successfully, (2) users can resolve conflicts with precision, and (3) users can compare objective functions over time and learn trade-offs between them. Thus, we validated CACTUS's success based on the following quantitative metrics:

\noindent \textbf{Task completion times: }
We measured task completion time when users were asked to:
(1) Report a conflict between a pair of objectives ($M=2.43 mins. \ [1.41, 3.45]$),
(2) Report highest conflict between three pre-defined objective functions ($M=5.12 mins. \ [2.81, 7.43]$). 
Next, we measured task completion time when participants: (1) Resolved conflicts between a pair of given objectives in an objective function ($M=3.03 mins. \ [2.59, 3.47]$), and (2) Resolved conflicts across the three objective functions ($M= 5.02 mins. \ [4.00, 6.04]$). 
The relatively lower task completion time (in comparison to writing code to perform the same task) answered \textbf{RQ1} that CACTUS makes it easy for users to successfully find and resolve conflicts.

\noindent \textbf{Correctness in conflict resolution: }
In addition, $13$ out of $14$ participants  successfully found the right conflicts between objectives.
Thus answering \textbf{RQ1} we observed $92.86 \%$ success rate among participants to find conflicts in an objective function. 
We also observed that every participant was able to successfully resolve conflicts ($100\%$ success rate). However, we found that in the case of two participants, while they successfully performed the interactions to resolve conflicts, the system failed to correctly record the specified changes and thus failed to update the objective function. This can be attributed to either: (1) a bug in the system, (2) a latency in the network, as the study was conducted completely online due to the on-going COVID-19 pandemic.
In future, we will look in to this issue further to understand the most likely cause.
However, given the high success rate of the task ($12/14 = 85.71 \%$), and the quick task completion time 
relative to writing programs or codes in Python or R, we consider these measures answer \textbf{RQ2}.

\noindent \textbf{Incremental comparison of objective functions: }
To answer \textbf{RQ3},
we further analysed log data to assess if participants were able to compare objective functions and learn trade-offs between objectives as they trained multiple classifiers. In doing so, we observed that on an average, participants iterated $10.23$ times ($M=10.23 \ [6.69, 13.77]$) to improve the given classifiers' baseline validation accuracy score of $78.24 \%$ on the Cancer mortality dataset \cite{cancerdata}.  $11$ out of the $14$ participants selected an objective function (to export, as their final selection) from a previous iteration in time. Furthermore, we observed a set of approaches to train new models: (1) Train a new classifier by changing weights only ($2/14$ participants), (2) Train a new classifier by resolving conflicts only ($4/14$ participants), and (3) a hybrid approach of the two which was the most popular ($8/14$ participants). 
We also found that $88.34 \%$ of the participants found success in improving the baseline classifiers' performance using these approaches.

	\begin{figure}[htbp]
\centering
		\includegraphics[width=\linewidth]{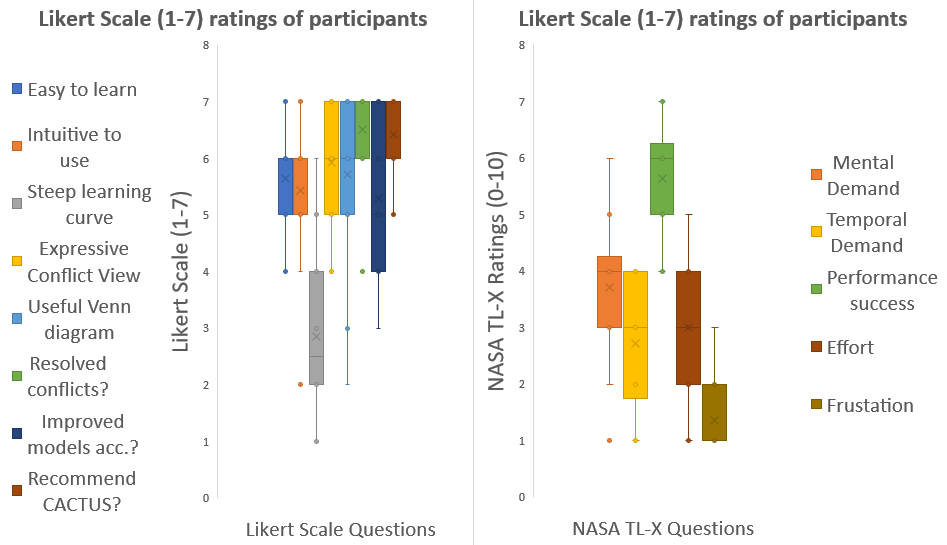}
		\caption{Results of Likert scale ratings and NASA TL-X scores for CACTUS.}
		\label{fig:workflow_tech}
		\vspace{-0.1cm}
	\end{figure}

\noindent \textbf{User preference ratings:}
We analysed Likert scale user preference ratings (on a scale of $1$ to $7$, higher is better) provided by users after they interacted with CACTUS.
Participants expressed that CACTUS was \textit{Easy to use} ($M=6.04 \ [5.85, 6.23]$), and it's interface was \textit{Intuitive} ($M=5.51 \ [5.05, 5.96]$). The majority also confirmed that the interface did not have a \textit{steep learning curve} for new users ($M=2.45 \ [1.25, 3.65]$), lower is better in this case).  Most of the participants felt that they were successful in resolving conflicts ($M=5.11 \ [4.95, 5.27]$), and they were able to incrementally improve models' validation accuracy score ($M=5.23 \ [5.01, 5.45]$).
Furthermore, the participants confirmed that the \textit{Conflict view} was expressive to not only help them find conflicts but also know which conflicts were more severe from the set ($M=6.03 \ [5.54, 6.52]$). Likewise the \textit{Venn diagram view} was found to be very useful to resolve conflicts ($M=5.32 \ [5.11, 5.53]$). However, majority of the participants found the \textit{Feature plots} were not as effective to resolve conflicts ($M=3.11 \ [1.55, 4.67]$).
Next, from the NASA-TLX survey, we observed that on average every participants' mental workload, and frustration towards the tasks were on the lower side ($M=2.11 \ [1.01, 3.21]$ out of a $10$ point scale; lower is better). 
Based on these findings, we are encouraged to continue research along the lines of conflict resolution for interactive objective functions. However, the qualitative feedback enlightened other aspects of the system that may need further improvement in the future. \looseness = -2

\subsection{Qualitative Analysis}

\noindent \textbf{Expressive visualisations and intuitive interactions:}
Most of the participants liked the visual representations used in CACTUS to represent conflict. In addition, their feedback confirmed that they liked the current interaction in the system to help them resolve conflicts. \textbf{P10} noted, \textit{``Easy to use, UI was straight forward. Resolving conflicts was easy. I liked that I could do things without much interruption and whenever I wanted the most.''}. However, \textbf{P07} pointed \textit{``I could easily find most severe conflicts, which was also my approach to resolve conflicts as I trained model. But, the sorting of the conflicts could be based on some rationale (eg. number of conflicts, or other criteria, etc.) to prioritise urgent conflicts first.''}. This feature is relevant and we plan to incorporate in the UI in future. \looseness = -2

\noindent \textbf{Incremental objective function exploration:}
We observed that participants enjoyed the workflow of incrementally changing objective functions to train better performing models. They either changed weights or they resolved most severe conflicts to create a variety of objective functions and trained models. \textbf{P04} expressed \textit{``I liked the plot on the top that shows you the history of the model performance, it was very helpful for me to keep track of the model improvement.''}. A few participants gamified the model training process, with the goal to train models that perform better than the models that were trained using weights recommended by the system. \textbf{P03} said \textit{``I liked the highlighted bar chart view with the attributes with the highest variance. That helped me tweak my weights as well as resolve conflicts, and in doing so I was trying to beat the machine to find a better [performing] model''}.

\noindent \textbf{Glean insights about the process:}
A few participants expressed the desire to learn a bit more about the process, in addition to be able to  interactively train models. For example, \textbf{P02} shared \textit{``At times it was hard to know how to change the weights and whether to move left or right when trying to resolve conflicts. I expect to see more visual cues on how to improve models' accuracy.''}. Similarly, \textbf{P01} noted \textit{``Doing things were easy, but doing them well took more background knowledge than I had. It would help to know how the function was driving the modeling process''}. In this iteration, we developed CACTUS as a proof of concept prototype that confirmed that conflicts can be detected and resolved interactively, however, in future, we plan to make this incremental modeling process more transparent using GUI elements. In the current setting, users can learn more about  the modeling process by exporting the objective function or the conflicts to the Jupyter notebook environment.

\noindent \textbf{Conflict resolution strategy:}
Broadly, we understood two main strategies participants used to resolve conflicts once they found which conflict to resolve. The first approach was to compare the shape of the data in conflict, to the data in the two objectives by observing the \textit{Feature plots}. Based on the shape resemblance they moved the conflicted data instances to one of the objectives. The other approach was to look at the \textbf{Variance bars} to find resemblance of the conflicted data to one of the objectives. When participants were not sure, how to resolve the conflict, they preferred to export it to Jupyter notebook to provide better examples by writing Python code. \textbf{P08} said, \textit{``My first step would be to find which variables are useful to analyze in the conflict view. I would then, look at the bar charts to see if the conflicts were closer to one of the objectives. If not, I would look at the scatterplot and make judgements by approximating the average location of each of the three groups.''}. In the future, we plan to provide more visualization supports to help users decide on how to resolve conflicts. \looseness  =-2

\subsection{Study Limitations}
In this study, the tasks are designed so that they can be completed within $60-70$ minutes. However, we understand that in real usage, these tasks can be more exploratory and may take longer times.
To construct classifiers we specified a hand chosen list of hyperparameters to the Optuna Auto-ML solver, with values that can be sampled from a specified domain range.
The study results may get affected if a different Auto-ML tool is selected or a different set of hyperparameters are tuned.
The scale of the data that we used are medium-sized ranging in thousands ($<50000$ samples). The results may vary if we use large datasets (e.g., $100k$ or more).
Due to the COVID-19 pandemic, we had to conduct the study completely remote, which had its own set of limitations in comparison to in-person controlled lab study settings such as lack of direct observations, network latency issues, in-adequate feedback from think-aloud protocols etc.
\section{Discussion and Limitations}

\noindent\textbf{Exploratory model space analysis: }
Enabling users to interactively construct multiple variants of objective functions empowers them to explore the model space (set of models defined by their unique hyperparameter settings) in an ad hoc iterative workflow. Users can adjust weights of objective, resolve conflicts, redefine objectives in Python and then import in CACTUS to visualise new objective functions. These functions trigger Auto-ML to search for models that are better aligned with their goals. The study further confirms this hypothesis that CACTUS encourages exploratory model space analysis using objective functions as a means to communicate user preferences to find optimal solutions to their ML task. Furthermore, we observed that visualising conflicts and model performance scores with respect to specific objectives, helps users interpret implication of the selected model on their expectation. In some way, we can assert that objective  functions support interpretation of models' performance, however, we understand more research needs to be done to confirm if interactive objective functions can further explain models.



\noindent \textbf{Collaborative objective functions and conflicts: }
From our experience with objective functions, we realised that in real world ML teams, objective functions are often designed by multiple members of the team to solve the same problem. In situations like this CACTUS can be a tool that helps compare multiple versions of objective functions visually, from different team mates, and then explore the space of possible objective functions and conflicts that may arise from each. That may open a new research area as an extension to our current work, in which users may like to resolve conflicts by collaboration across different geo-locations, or from different devices (e.g, from browsers accessing CACTUS synchronously and resolving conflicts). We plan to look into this area of work in the near future. \looseness = -2

\noindent \textbf{Gamification of modeling using objective functions: }
In the study we observed a few participants, gamified the process of model construction, by testing various weight settings of objectives, resolving conflicts, or specifying new objectives. Their goal was to beat the previous models' performance score and also to perform better than the model constructed using system recommended weights. 
Many participants elicited there is a lot of value in this, as not only it helps them ideate and explore faster, but also makes the process of finding a suitable model more fun. We realise incremental construction of objective functions, and being able to revert back to any time step in this process empowers them to freely explore and experiment with their hypothesis. In future, we see lot of potential in supporting the interface with features that further encourage and guide users in this process of gamified model creation. \looseness = -2

\noindent \textbf{Current limitations in conflict resolution: }
Though CACTUS allows users to incrementally create and compare objective functions in the process of resolving conflicts, through the study we found many aspect of our work that needs further research.
For example, we found the current interface showing the incremental view of objective functions (\textit{Model spark bars} and the \textit{Objective function gallery}) is often difficult to track because it only supports sequential record. 
Every change in one parameter leads to a model retraing, and saves a new copy of the objective function. This may be difficult to track when users have iterated many times. We aspire to research further to find potentially better design choices in-terms of visual design and interaction to present the timeline representation of models and objective functions.

Another issue, we realise is that data exploration is critical to resolve conflicts and design good objective functions. A few participants in the study, who were less versed with the data, felt very difficult to know how to change the weights and whether to move left or right when trying to resolve conflicts. Sometimes this workflow felt like a trial and error rrocess to them. 
In CACTUS we focused on conflict resolution and objective function comparison, while separated the task of data exploration in Jupyter notebook. While that may work for advanced users with Python and Jupyter notebook experience, it seemed for intermediate users, a part of the interface should support data exploration in the future.

\noindent \textbf{Scalable conflict detection: }
The current prototype's implementation is designed to be model agnostic, meaning it can work with classification, or regression models (supervised ML). It can also work with sequence data such as text, time-series and image data. Currently, the interactions are tested on medium sized data sets (i.e., with data samples in the range of 100k).
The design of the visualisation uses Phaser JS (a 2D game engine library for the web) with WebGL on canvas implementation.
In the next iteration the conflicts can be ordered, grouped and can be extended to represent more than two objectives per row in the table. Though we tested with multiple conflicts per row, we discarded the approach, as it may lead to ineffective selection of conflict intersections in the venn diagram view. In future, we wish to research potential solution for multi-objective venn diagram based solutions to represent conflicts.

\section{Conclusion}
Through this work, we present our research on making interactive visualizations of objective functions in the process of constructing classifiers. Furthermore, with our visual analytic system CACTUS, we demonstrated a novel technique that interactively detects and resolves conflicts among objectives in an multi-objective objective function. In this paper, we also discussed a list of various types of conflicts that may occur when users interactively specify objective functions to Auto-ML model solvers to classify tabular data. With a quantitative and qualitative user study we show that our technique helps users to interactively detect and resolve conflicts between objectives, and incrementally train ML classifiers in tandem with a Jupyter notebook environment.
In future, we are motivated to continue our research on interactive objective functions to guide users explore and ideate multiple variants of these functions as they collaboratively construct ML models. \looseness = -2

\section{Acknowledgements}
Support for the research is provided by DARPA FA8750-17-2-0107. The views and conclusions contained in this document are those of the authors and should not be interpreted as representing the official policies, either expressed or implied, of the U.S. Government.\looseness=-1

\ifCLASSOPTIONcaptionsoff
  \newpage
\fi



\bibliographystyle{IEEEtran}
\bibliography{IEEEabrv,template,templ2,newrefs}
%




%

\begin{IEEEbiographynophoto}{Subhajit Das}
is a Ph.D. Computer Science student at the School of Interactive Computing, Georgia Institute of Technology, USA. His research interests include interactive machine learning, model optimization/selection, and designing human-in-the-loop based visual analytic systems. 
\end{IEEEbiographynophoto}

\begin{IEEEbiographynophoto}{Alex Endert}
is an Assistant Professor in the School of Interactive Computing at Georgia Tech. He directs the Visual Analytics Lab, where him and his students explore novel user interaction techniques for visual analytics. His lab often applies these fundamental advances to domains including text analysis, intelligence analysis, cyber security, decision making, and others. He received his Ph.D. in Computer Science at Virginia Tech in 2012.
\end{IEEEbiographynophoto}




\end{document}